\documentclass[conference]{IEEEtran}
\IEEEoverridecommandlockouts

\usepackage{cite}
\usepackage{amsmath,amssymb,amsfonts}
\usepackage{algorithmic}
\usepackage{graphicx}
\usepackage{textcomp}
\usepackage{xcolor}

\usepackage{acro}
\usepackage{xcolor}
\usepackage{subcaption}
\usepackage{algorithm}
\usepackage{algorithmic}
\usepackage{multirow}
\usepackage{adjustbox} 
\usepackage{threeparttable}
\usepackage{tabularx}
\usepackage{pifont}
\usepackage[colorlinks=true, allcolors=blue]{hyperref}

\newcommand{\cmark}{\ding{51}}   
\newcommand{\xmark}{\ding{55}}   

\newcommand{\adrianfeed}[1]{\textcolor{black!70!black}{#1}}
\newcommand{\manuelfeed}[1]{\textcolor{black!70!black}{#1}}
\newcommand{\danielfeed}[1]{\textcolor{black!70!black}{#1}}
\newcommand{\christianfeed}[1]{\textcolor{black!70!black}{#1}}

\newif\ifanon

\ifanon
\newcommand{\repo}{[anonymised for review]}
\newcommand{\makeauthors}{\author{Anonymous Authors}}
\newcommand{\makeack}{Anonymised for review.}
\else
\newcommand{\repo}{\url{https://github.com/enlite-ai/maze-flatland}}

\newcommand{\makeauthors}{\IEEEauthorblockN{
		Alberto Castagna\IEEEauthorrefmark{1}\,
		Stefan Zahlner\IEEEauthorrefmark{1}\,
		Adrian Egli\IEEEauthorrefmark{2}\,
		Christian Eichenberger \IEEEauthorrefmark{3}\\
		Daniel Boos\IEEEauthorrefmark{2}\,
		Manuel Meyer\IEEEauthorrefmark{3} \
		Anton Fuxjäger\IEEEauthorrefmark{1}\,
	}

\IEEEauthorblockA{
	\IEEEauthorrefmark{1}enliteAI, AT \quad
	\IEEEauthorrefmark{2}SBB CFF FFS, CH \quad
	\IEEEauthorrefmark{3}Flatland Association, CH
}

\IEEEauthorblockA{
	\textit{Corresponding author:} a.fuxjaeger@enlite.ai
}
}

\newcommand{\makeack}{\danielfeed{The authors acknowledge the AI4REALNET project, funded by the European Union’s Horizon Europe Research and Innovation Programme under Grant Agreement No.~101119527, and the Swiss State Secretariat for Education, Research and Innovation (SERI). The views and opinions expressed are solely those of the authors and do not necessarily reflect those of the European Union or SERI; neither the EU nor the granting authority can be held responsible for them.}}
\fi

\IfFileExists{acronyms.tex}{\DeclareAcronym{vrsp}{
  short = VRSP,
  long  = Vehicle Routing and Scheduling Problem
}
\DeclareAcronym{pomdp}{
  short = POMDP,
  long  = Partially Observable Markov Decision Process~\cite{kaelbling1998planning}
}

\DeclareAcronym{rl}{
  short = RL,
  long  = Reinforcement Learning
}

\DeclareAcronym{sl}{
  short = SL,
  long  = Supervised Learning
}

\DeclareAcronym{marl}{
  short = MARL,
  long  = Multi-Agent Reinforcement Learning
}

\DeclareAcronym{or}{
  short = OR,
  long  = Operational Research
}
\DeclareAcronym{mads}{
  short = \textbf{MADS},
  long  = Multi-Agent Departure Scheduling 
}
\DeclareAcronym{mapf}{
  short = \textbf{MAPF},
  long  = Multi-Agent Path Finding
}

\DeclareAcronym{pp}{
short = PP,
long= Prioritized Planning
}

\DeclareAcronym{lns}{
short = LNS,
long= Large Neighbourhood Search
}

\DeclareAcronym{mcp}{
short = MCP,
long= Minimum Communication Policy
}

\DeclareAcronym{lcppo}{
short = LCPPO,
long= Local Critic Proximal Policy Optimization
}

\DeclareAcronym{ctde}{
short = CTDE,
long= Centralized-Training-Decentralized-Execution
}

\DeclareAcronym{dp}{
short = DP,
long= Decision Point
}

\DeclareAcronym{mdp}{
short = MDP,
long= Markov Decision Process
}
\DeclareAcronym{bc}{
short = BC,
long= Behavioural Cloning
}

\DeclareAcronym{mcts}{
short= MCTS,
long= Monte Carlo Tree Search
}

\DeclareAcronym{ai}{
short= AI,
long= Artificial Intelligence
}}{}

\def\BibTeX{{\rm B\kern-.05em{\sc i\kern-.025em b}\kern-.08em
    T\kern-.1667em\lower.7ex\hbox{E}\kern-.125emX}}
\begin{document}

\title{Towards Autonomous Railway Operations: A Semi-Hierarchical Deep Reinforcement Learning Approach to the Vehicle Rescheduling Problem}

\author{\makeauthors}

\maketitle

\begin{abstract}
	
Managing disruptions in railway traffic management is a major challenge. Rising traffic density and infrastructure limits increase complexity, making the \ac{vrsp} difficult to solve reliably and in real time. While \ac{or} methods are widely used, most dispatching still relies on human expertise due to the problem’s exponential combinatorial complexity. \ac{rl} has gained attention for its potential in multi-agent coordination, but existing \ac{rl} approaches often underperform \ac{or} methods and struggle to scale in dense rail networks. 

This paper addresses this gap from a machine learning perspective by introducing a semi-hierarchical \ac{rl} formulation tailored to operational railway constraints. The method separates dispatching from routing through dedicated action and observation spaces, enabling policies to specialise in distinct decision scopes and addressing the imbalance between rare dispatch decisions and frequent routing updates. The approach is evaluated on the Flatland-RL simulator across five difficulty levels and 50 random seeds, with 7 to 80 trains. Results show substantially improved coordination, resource utilisation, and robustness compared with heuristic baselines and monolithic \ac{rl}, nearly doubling the number of trains reaching their destinations, while keeping deadlock rates below 5\% and adaptively sequencing, delaying, or cancelling trains under heavy congestion.

\end{abstract}


\section{Introduction}

Planning and rescheduling train runs for ever-increasing traffic volumes is a complex task requiring months of expert preparation. During operations, specialists in operations centres monitor traffic flow and respond to disruptions in near real time, but dispatchers must make rapid decisions within short horizons. With rising traffic density and demand for flexibility, current dispatching processes and methods are reaching their limits, leaving dispatchers little time to react.

In systems with more than 10,000 daily journeys, minor disruptions significantly impact service quality and network stability, making dynamic rescheduling as critical as initial timetable planning.

The challenge lies in developing scalable algorithms that shift focus from local dispatch areas to global network solutions. System stability depends on shrinking time buffers, making global optimisation essential since any local adjustment affects the entire network. \ac{ai}, particularly \ac{marl}, offers promising approaches for real-time dispatching and rescheduling. However, scalability and reliability questions in real-world environments remain unsolved.

We study these questions using Flatland-RL, an open-source railway simulator modeling multi-agent train movement with discrete time and cell-level occupancy. The environment captures core railway characteristics including shared infrastructure, bottlenecks, dispatch timing, and routing conflicts. Recent work in this environment shows that existing \ac{rl} methods remain limited by monolithic decision structures, insufficient dispatch exploration, and poor scalability in dense traffic, motivating architectures that separate decision scopes and reflect real operational structures. This motivates the need for architectures that separate decision scopes and better reflect the operational structure of real rail systems.

To address these limitations, this work introduces \textit{Maze-Flatland}, a semi-hierarchical multi-agent \ac{rl} framework for the \ac{vrsp} in dense rail networks. Unlike monolithic policies, Maze-Flatland separates control into two coordinated levels: dispatching (\ac{mads}) and routing (\ac{mapf}). This decomposition reduces task interference and exploration imbalance, improving scalability, coordination, and learning stability. \adrianfeed{Its performance is benchmarked against heuristic baselines, including \ac{pp}~\cite{laurent2021flatland}, and learning-based methods such as TreeLSTM~\cite{jiang2022multi}, showing substantial gains in coordination efficiency and deadlock reduction.} A train is considered deadlocked if it can not move for rest of the episode.

The main contributions are: (i) analysis of \ac{rl} approaches in Flatland-RL highlighting scalability limits; (ii) a novel semi-hierarchical two-phase \ac{rl} approach for \ac{vrsp} in railway; and (iii) demonstration that this solution scales effectively with traffic density, enhancing performance and robustness.

The remainder of this paper is organised as follows. Section~\ref{sec:problem_formulation} introduces the \ac{vrsp} in railway operations. Section~\ref{sec:rwork} reviews state-of-the-art approaches. Section~\ref{sec:motivation} analyses agent behaviour and motivates semi-hierarchical design. Section~\ref{sec:our_approach} presents the proposed framework with control and observation spaces. Section~\ref{sec:sim_setup} describes simulation setup and metrics. Section~\ref{sec:results} reports results, Section~\ref{sec:ablation} provides ablation analysis, and Section~\ref{sec:conclusion} summarises contributions and future directions.

\section{Problem Formulation}
\label{sec:problem_formulation}

\christianfeed{We consider the \ac{vrsp} in railway operations as the real-time process of rescheduling train movements in response to disruptions~\cite{li2007vehicle} such as delays, failures, or resource shortages}. \adrianfeed{When the current state diverges from the initial plan, replanning is triggered and must be performed within a short time window}~\cite{lindenmaier2022efficient} to selectively revise affected train schedules while minimising the impact on \manuelfeed{the system}.

\danielfeed{This study focuses exclusively on operational disruptions such as train malfunctions, limiting the problem scope by assuming no shortages of staff or rolling stock.} Although real-world operations involve multiple optimisation objectives, \manuelfeed{here we prioritise the arrival ratio of the available trains while maintaining deadlock-free operation. Minimising total delay is desirable but not the primary target}.

The \manuelfeed{procedure in real-world railway operations} is twofold: first, operators generate an optimal baseline timetable and allocate resources for all train categories over a shared time window; second, they must adapt this timetable in real time by reassigning routes and resources whenever operational disruptions occur. \christianfeed{In this work, we specifically address the second phase, which is triggered when disruptions require immediate timetable and resource adjustments to maintain system service.}

\danielfeed{While many current traffic management operations distribute control among human operators responsible for different sectors or areas,} \manuelfeed{fully automated} solutions rely on decentralised control distributed at the train level, which ensures scalability across network configurations and fleet sizes.

In this work, each train is managed by a single agent, and the two terms are used interchangeably. Each agent perceives a partial, agent-centered observation based on its planned path and forecasted state. \christianfeed{This prediction may not accurately reflect future conditions, because in a distributed control setting, other trains execute independent actions that can alter the shared environment. As a result, an individual agent's anticipated future position is based only on its own estimation and local information, which may diverge from the actual state if other agents behave unexpectedly or if unforeseen interactions occur.} Finally, we do not limit communication range nor consider communication, privacy, or safety issues.

The problem is modelled as a multi-agent system in which several agents operate within a shared environment under partial observability. Although each agent aims to complete its journey on time, they share limited infrastructure and may compete for the same resources, creating local conflicts. Their individual objectives are independent, but their behaviours are interdependent, since the actions of one agent can constrain or delay others. As a result, effective solutions require coordinated collaboration~\cite{ferber1999multi}.

\section{Related Work}
\label{sec:rwork}

\christianfeed{Existing \ac{vrsp} solutions can broadly be divided into two classes: \textit{\ac{or}}-based optimisation and \textit{\ac{rl}}-based approaches. \ac{or} methods encompass both mathematical optimisation models, which formally encode objectives and constraints, and heuristics, which are typically based on domain-specific rules or simple decision strategies, and are relatively easy to implement but difficult to adapt, since operational changes require manual redesign and scalability degrades in real-time settings.}

\begin{table*}[htb!]
	\caption{Summary of related work.}
	\label{tab:related_work}
	\centering
	\begin{threeparttable}
		\begin{adjustbox}{max width=\textwidth} 
			\begin{tabularx}{.83\textwidth}{@{}
					|l|c|c|c|c|c|X|
					@{}}
				
				\hline 
				
				\multirow{2}{*}{\textbf{Reference}}
				& \multicolumn{2}{c|}{\multirow{1}{*}{\textbf{Approach}}}
				& \multirow{2}{*}{\textbf{\adrianfeed{Novelty}}}
				& \multirow{1}{*}{\textbf{Local}}
				& \textbf{\adrianfeed{Rail}}
				& \multicolumn{1}{c|}{\multirow{2}{*}{\textbf{Limitations}}} \\
				
				&  \textbf{Disp} & \textbf{Rout}  &  & \multirow{1}{*}{\textbf{Obs}} & \textbf{\adrianfeed{Topology}} &  \\
				\hline
				
				\textit{An\_Old\_Driver} & \multicolumn{2}{c|}{\multirow{2}{*}{\acs{or}}} & \acs{pp}, \acs{lns} & \multirow{2}{*}{\xmark} & \multirow{2}{*}{\cmark} & \multirow{2}{*}{centralised and iterative approach}\\ 
				\multirow{1}{*}{2021 \cite{li2021scalable}} & \multicolumn{2}{c|}{} &partial re-planning & & & \\ \hline

				\multirow{1}{*}{\textit{JBR\_HSE}} & \multicolumn{1}{c|}{\multirow{2}{*}{\acs{sl}}} & \multirow{2}{*}{\ac{rl}} & agents share messages & \multirow{2}{*}{\cmark} & \multirow{2}{*}{\cmark} & \multirow{2}{*}{scheduler with fixed threshold}\\ 
				2021~\cite{laurent2021flatland} & & & supervised dispatching & & & \\\hline
				
				\multirow{1}{*}{\textit{Netcetera}} & \multicolumn{2}{c|}{\multirow{2}{*}{\ac{rl}}} & \multirow{1}{*}{rail-as-graph} & \multirow{2}{*}{\cmark} & \multirow{2}{*}{\cmark} & \multirow{2}{*}{under-usage of resources}\\ 
				2021~\cite{laurent2021flatland} & \multicolumn{2}{c|}{} & \multirow{1}{*}{priority scheduling} & & & \\\hline

				\multirow{1}{*}{\textit{TreeLSTM}} & \multicolumn{2}{c|}{\multirow{2}{*}{\ac{rl}}} & 
				\multirow{1}{*}{path encoding} & \multirow{2}{*}{\cmark} & \multirow{2}{*}{\cmark} & \multirow{2}{*}{under-usage of resources}\\ 
				2022~\cite{jiang2022multi} &\multicolumn{2}{c|}{} & \multirow{1}{*}{ with TreeLSTM} & & & \\\hline
				
				\multirow{1}{*}{\textit{\acs{lcppo}}} & \multicolumn{2}{c|}{\multirow{2}{*}{\ac{rl}}} & 
				\multirow{1}{*}{dynamic local} & \multirow{2}{*}{\cmark} & \multirow{2}{*}{\cmark} & \multirow{2}{*}{poor scalability and frequent deadlocks}\\ 
				2024~\cite{zhang2024improving} &\multicolumn{2}{c|}{} & \multirow{1}{*}{per-group critic} & & & \\\hline

				\multirow{1}{*}{\textit{Schneider et al.}} & \multicolumn{2}{c|}{\multirow{2}{*}{\ac{rl}}} & 
				\multirow{1}{*}{high-fidelity} & \multirow{2}{*}{\cmark} & \multirow{2}{*}{\xmark*} & \multirow{2}{*}{simulator-dependent and fixed rail layout}\\ 
				2024~\cite{schneider2024intelligent} &\multicolumn{2}{c|}{} & \multirow{1}{*}{proprietary simulator} & & & \\\hline
				
				\multirow{2}{*}{\textbf{Our solution}} & \multicolumn{1}{c|}{\multirow{2}{*}{\textbf{\ac{rl}}}} & \multicolumn{1}{c|}{\multirow{2}{*}{\textbf{\ac{rl}}}} & \multirow{1}{*}{\textbf{semi-hierarchical}} &\multirow{2}{*}{\cmark} & \multirow{2}{*}{\cmark} & \multirow{2}{*}{\textbf{constant speed}} \\
				& & &\multirow{1}{*}{\textbf{decision making}} & & & \\ 
				\hline 
			\end{tabularx}
		\end{adjustbox}
		\begin{tablenotes}
			\small
			\item *Evaluated on different sections of the same network.
		\end{tablenotes}
	\end{threeparttable}
\end{table*}

\ac{rl} methods learn adaptive policies from interaction rather than rules, enabling broader generalisation. In multi-agent settings they gain scalability through decentralised decision-making. Despite this flexibility, deploying \ac{rl} in distributed systems remains challenging, requiring careful reward design and coordination among agents.

Flatland-RL challenges have served as a benchmark for studying the \ac{vrsp} in large-scale railway environments. In the 2020 edition, \textit{An\_Old\_Driver}~\cite{li2021scalable} combined \ac{pp}~\cite{silver2005cooperative}, \ac{lns}~\cite{shaw1998using}, and \ac{mcp}~\cite{ma2017multi} for deadlock-free re-planning. The \textit{JBR\_HSE} team~\cite{laurent2021flatland} proposed an \ac{rl} solution using tree-based observations and neighbour communication, and introduced a \textit{scheduler} via \ac{sl} to dispatch trains based on estimated success probability. \textit{Netcetera}~\cite{laurent2021flatland} adopted graph-based observations but reported reduced resource utilisation and noise from deeper lookahead.

Building on these ideas,~\cite{jiang2022multi} introduced a TreeLSTM model to encode future paths and anticipate deadlocks, improving toward \ac{or} baselines but underusing capacity in low-density settings. More recently,~\cite{zhang2024improving} introduced \ac{lcppo}, a PPO extension with local critics. Although it improved success rates over earlier \ac{rl} baselines, its scalability remains limited.

\ac{marl} shows promise for realistic railway scheduling~\cite{schneider2024intelligent}, though coordination, observation design and dispatch control remain open challenges.


Most existing methods rely on heuristic optimisation or monolithic \ac{rl} models that struggle with scalability and adaptability. To overcome these limitations, this work formulates the \ac{vrsp} as a semi-hierarchical \ac{marl} problem in which a dispatching policy decides which trains to release and when, and  a routing policy handles on-track path following. This decomposition captures the operational distinction between dispatching trains and routing them once active, and it forms the foundation of the Maze-Flatland framework.

\christianfeed{Table~\ref{tab:related_work} summarises key approaches to the \ac{vrsp} in railway operations, comparing them by \textit{Approach} (heuristic or \ac{rl}-based). To provide a more precise breakdown, we distinguish two functional dimensions: \textit{Dispatching} (disp), which governs when trains are released and how their sequencing or priorities are managed, and \textit{Routing} (rout), which concerns how a method determines a train’s path through the network.}
The table also reports \adrianfeed{\textit{Novelty} (the central contribution of each method)}, \textit{Local Obs.} (whether decisions rely on localised agent observations), \adrianfeed{\textit{Rail Topology} (fixed layouts versus randomly generated networks)}, and \textit{Limitations} (the method’s main weaknesses).

Unlike previous monolithic \ac{rl} methods, Maze-Flatland decomposes decision-making into two specialised models, strengthening the learning signal through subtask-specific observations and reducing noise propagation. To our knowledge, no prior Flatland-RL work learns separate dispatching and routing policies as \ac{marl} agents with distinct observation spaces, although earlier work has explored heuristic dispatch–routing decompositions~\cite{laurent2021flatland}.

\section{Motivation}
\label{sec:motivation}
\adrianfeed{In Flatland-RL, at each timestep each train selects one of five actions: \textit{NOOP} (maintain current state), \textit{STOP} (hold the train in the current cell), or a movement action: \textit{FW} (forward), \textit{LEFT}, or \textit{RIGHT} (turn).}

\begin{figure}[htbp!]
	\centering
	\includegraphics[width=.85\columnwidth]{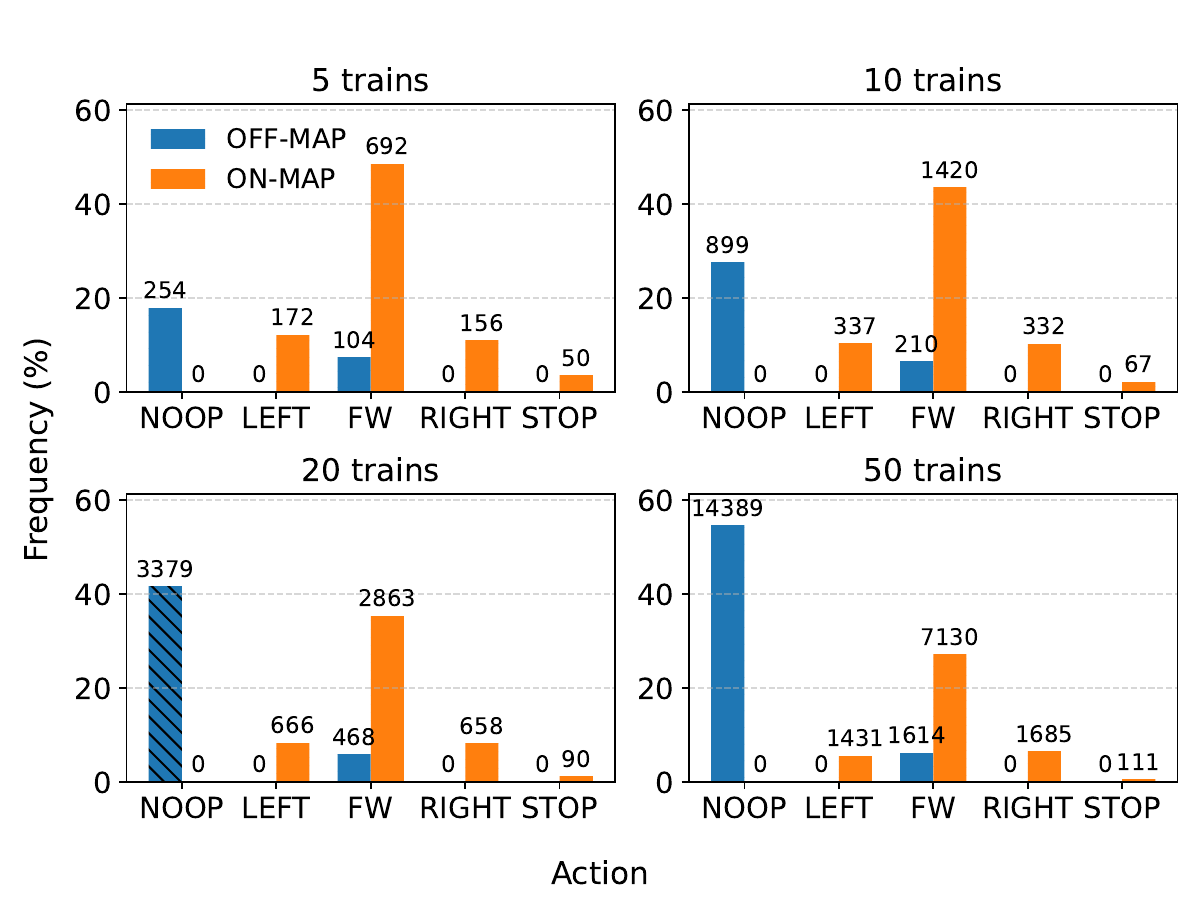}
	\caption{Action distribution across 20 episodes for a trained agent on random maps where at least 90\% of trains arrive \manuelfeed{at} their target.}
	\label{fig:action_distribution}
\end{figure}

During early experimentation, our analysis of agent behaviour in the environment revealed a strong imbalance between \textit{OFF-MAP} (undispatched) and \textit{ON-MAP} (active) experiences, as shown in Figure~\ref{fig:action_distribution}. Since dispatching is implicitly tied to the forward action, fewer than 15\% of these actions correspond to departures. At higher densities, agents increasingly choose the \textit{NOOP} to delay dispatch under congestion.

The observed behavioural imbalance reflects a well-known limitation of monolithic \ac{rl} policies: frequent subtasks dominate learning signals, while rare but high-impact decisions (dispatching) remain underexplored. 
This problem is known as the Hard-Exploration Problem~\cite{ecoffet1901go}, which arises in environments with very sparse or even deceptive rewards. Random exploration is prone to failure as it will rarely discover successful states or obtain meaningful feedback from the environment. Hierarchical RL theory therefore motivates explicit modelling of rare, temporally extended decisions as separate high-level policies to stabilise exploration and value estimation. In railway operations, dispatch timing is precisely such a rare yet high-impact decision. The absence of an explicit dispatch action further reduces its exploration, reinforcing a routing-dominant bias. Our experiments indicate that this effect becomes more pronounced when using incremental training approaches, as in~\cite{laurent2021flatland, jiang2022multi}. In addition, dispatching and routing pull the policy in different directions: delaying a train may support global flow, whereas locally it appears suboptimal. A single value function must reconcile these competing signals, which complicates the learning process. These factors motivate separating the decision scopes. The semi-hierarchical framework introduced in the next section defines distinct state and action spaces for dispatching and routing, reducing interference and enabling task-specific observations.


\section{\manuelfeed{Method}}
\label{sec:our_approach}

To address the \ac{vrsp}, we adopt a semi-hierarchical \ac{marl} formulation that separates dispatching from routing. Our design follows hierarchical \ac{rl} principles, decomposing the problem into subgoals and subtasks. This aligns with established hierarchical \ac{rl} frameworks such as MAXQ~\cite{dietterich2000hierarchical}, which assume that the designer specifies useful subgoals and corresponding subtasks. In contrast to fully hierarchical \ac{rl}, however, our method optimizes a single global objective and enforces a strict temporal ordering between subpolicies: the dispatching policy is executed first, and the routing policy is used only after dispatching has been completed, rather than being invoked arbitrarily at any time. By imposing this structure, the designer restricts the space of admissible policies during reinforcement learning, which typically improves tractability and sample efficiency.

We model the \ac{vrsp} as a \ac{pomdp} with partial observability limiting agents to local information.  The environment is partitioned into dispatching and routing subspaces $(S_d, S_r)$ with action sets $(A_d, A_r)$. Transitions $T$ and rewards $R$ follow standard \ac{mdp} definitions. An agent starts in dispatching space $(S_d, A_d)$ and transitions to routing space $(S_r, A_r)$ after a dispatch action.

Dispatching and routing operate on different temporal and spatial scales. Dispatching occurs only once per train and determines when it enters the network, giving it a sparse but high-impact role. Routing, in contrast, involves frequent local decisions made throughout the journey. In a \ac{pomdp}, combining them into a single policy forces the agent to learn a joint value function over heterogeneous state features, creating high variance and action imbalance. These structural differences motivate a decomposition in which each policy focuses on its own decision scope. \manuelfeed{Based on this formulation, we design two tailored action–observation spaces in Flatland-RL that reflect the requirements of dispatching and routing respectively.}

\begin{figure}[b!]
	\centering
	\includegraphics[width=.72\columnwidth]{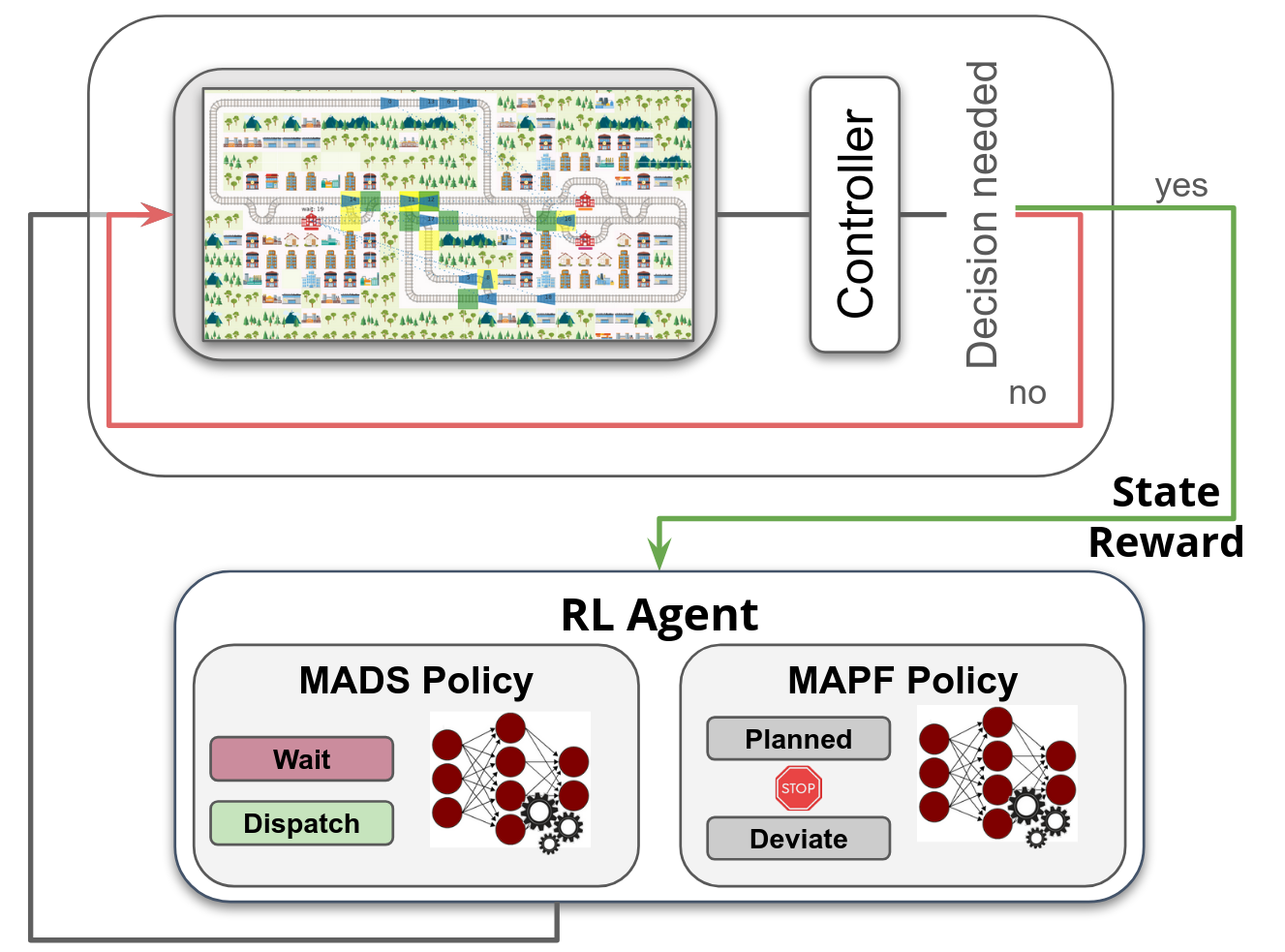}
	\caption{Semi-hierarchical control loop with decision controller for skipping.}
	\label{fig:semi-hierarchical-control-loop}
\end{figure}

The semi-hierarchical control loop, illustrated in Figure~\ref{fig:semi-hierarchical-control-loop}, governs both dispatching and routing. The environment provides each agent with an observation, which is processed by either \ac{mads} for dispatching or \ac{mapf} for routing. The decision controller triggers decisions only when necessary, such as at switches.

\ac{mads} — The dispatching agent selects between two actions: \textit{Dispatch} or \textit{Wait}. The \textit{Wait} action delays the decision by one timestep, while \textit{Dispatch} places the train onto the rail, enabling the use of the routing policy.  \ac{mads} learns a decentralised policy based on current network conditions.

The observation combines global metrics, such as total trains, map occupancy, episode length, and remaining time, with conflict information assessing whether the network can accommodate another train. At dispatch time, active trains share their top-$k$ paths, which are compared with those of the waiting agent to compute overlapping cells (\textit{N\_conflict\_cells}) and distances to potential clashes (\textit{Dist\_conflict}) as reported in Figure~\ref{fig:agent-architecture}.

\ac{mapf} — For routing, the action space reflects real-world constraints. At each switch, a train may face up to two possible directions and can follow the \textit{Planned} path, \textit{Deviate} to an alternative route, or \textit{Stop} until a new movement signal is given. Deviations trigger replanning along the shortest path to the target, ignoring obstacles. Each agent follows a predefined route until deviation occurs, mirroring real-world operations where dispatchers assign fixed paths.

The observation refines the tree-based structure (\textit{TreeObsForRailEnv}) from the Flatland-RL repository~\cite{flatland-git}. Cells up to the next \adrianfeed{\ac{dp}, defined as cells where an on-map train has one or more routing options,} are collapsed; each path is encoded using a binary representation distinguishing the \textit{shortest} and \textit{alternative} paths. For branches beyond the first decision point, only deadlock presence, distance, and travel cost are recorded. 

\begin{figure}[h]
	\centering
	\includegraphics[width=0.55\columnwidth]{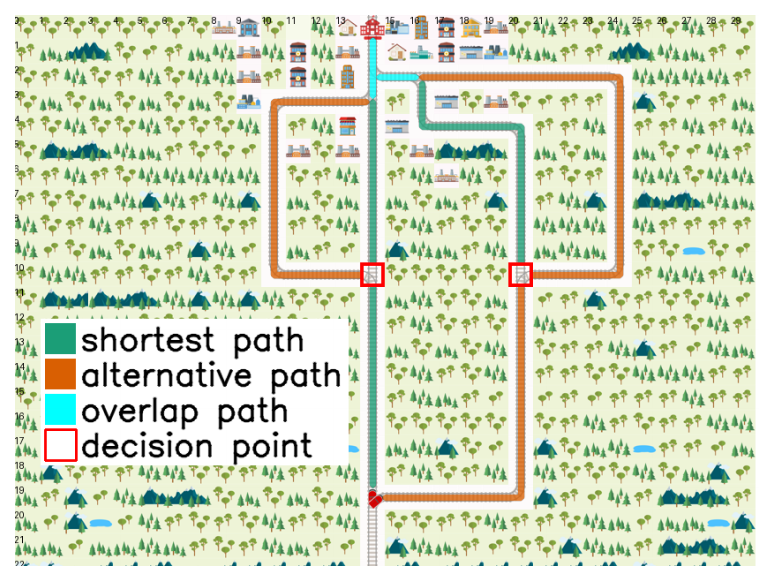}
	\caption{\ac{mapf} — representation of an observation.}
	\label{fig:mapf-obs}
\end{figure}

To enable decentralised coordination, each agent also observes the observation of the first encountered agent along its projected path at depth 1, where the projected path includes both the shortest and the alternative branch. \adrianfeed{The observation is depicted in Fig.~\ref{fig:mapf-obs}.}

Each agent is equipped with a \textbf{decision controller} that prevents unnecessary decisions ~\cite{roost2020improving} under masking conditions. For \ac{mads}, actions are skipped if the train has not yet reached its departure time or if no valid path exists to the target within the episode horizon. For \ac{mapf}, masking limits available actions to those feasible at the current cell, considering route divergence and nearby conflicts.



\begin{algorithm}[h!]
	\small	
	\caption{Maze-Flatland decision process: Semi-hierarchical agent–environment interaction.}
	\label{alg:semi_hierarchical_rl}
	\begin{algorithmic}[1]
		\STATE initialise environment~$E$
		\STATE initialise observation builders for dispatching~($Obs_{disp}$) and routing~($Obs_{rout}$)
		\STATE initialise policies \acs{mads} ($\pi_{mads}$) and \acs{mapf} ($\pi_{mapf}$)
		\FOR{each timestep $t_j$}
		\STATE Compute $\mathcal{A}_t \gets \langle a_t^0, a_t^1, \dots, a_t^n \rangle$
		\FOR{each agent $\alpha^i$}
		\IF{$\alpha^i$ is outside of the map}
		\STATE $s_t^i \xleftarrow{} Obs_{disp}(E)$ 
		\STATE $a^i \xleftarrow{} \pi_{mads}(s_t^i)$
		\ELSE
		\STATE $s_t^i \xleftarrow{} Obs_{rout}(E)$
		\STATE $a^i \xleftarrow{} \pi_{mapf}(s_t^i)$
		\ENDIF
		\ENDFOR
		\STATE step $E$ with $\mathcal{A}_i$
		\ENDFOR
	\end{algorithmic}
\end{algorithm}

The semi-hierarchical \ac{marl} framework is outlined in Alg.~\ref{alg:semi_hierarchical_rl}. The environment and the observation spaces for dispatching ($Obs_{disp}$) and routing ($Obs_{rout}$) are initialised in lines~1–2, followed by policy initialisation at line~3. For each timestep $t_i$ in an episode, the joint action vector $\mathcal{A}_i$ is computed from the two policies. If an agent has not yet departed, its observation from $Obs_{disp}$ is passed to \ac{mads} to select an action (lines~7–9); otherwise, $Obs_{rout}$ and \ac{mapf} are used (lines~10–12). Finally, the environment steps with $\mathcal{A}_i$~(line~15).

The semi-hierarchical decomposition isolates two decision layers with different scopes: dispatching controls global traffic density, whereas routing handles local conflict resolution by selecting deviations or stops.

\section{Simulation Setup}
\label{sec:sim_setup}

All experiments were conducted on the Flatland-RL environment~\cite{flatland-git}, using versions $4.0.4$ and $3.0.15$ (the latter for backward compatibility with existing baseline implementations). Aside from the random-disruption generator, the two versions are identical; we therefore used the same timetables, rail configurations, and disruption profiles in all experiments, varying only the sampled malfunction timings.

We adopt a high-concurrency evaluation regime. To create this setting, we run all trains at a constant speed of~1, which causes many trains to be simultaneously active on the network. This produces denser traffic, more frequent conflicts, and a substantially harder dispatch-routing problem. Fractional-speed profiles, in contrast, slow down part of the fleet and effectively stretch the operational window, lowering concurrent activity and reducing congestion. As our goal is to assess scalability under dense traffic, all baselines and Maze-Flatland are evaluated using the constant-speed configuration.

To support the semi-hierarchical decision structure, we built a Flatland-RL wrapper enabling action masking and step-skipping. Implementation is open-source.\footnote{\repo}

Agents exchange observations under the assumption of flawless communication; both communication failures and adversarial messaging are out of scope. For the \ac{mapf} policy, we limit the observation depth to 2 (full segment after the next \ac{dp}), as shown in Figure~\ref{fig:mapf-obs}. Increasing this depth would \adrianfeed{raise computation time while adding superfluous and inaccurate information, since observation relies on projected rather than actual future train positions}. 

We evaluated Maze-Flatland across multiple instances to assess the robustness of the algorithm under increasing complexity and denser schedules. Inspired by past Flatland-RL challenges, we \manuelfeed{ran} our agent over 50 seeds in \manuelfeed{a} setup similar to the competition setting~\cite{laurent2021flatland}. \manuelfeed{We simulated five levels with an increasing number of trains, ranging from 7 to 80}.

During evaluation, all trains move at a constant speed of 1, traversing one cell per timestep, and malfunctions last between 20 and 50 timesteps. The maximum number of rails between cities and within each city is fixed at 2. The number of trains, map size, number of cities, and malfunction probability for each test level are given in Table~\ref{tab:challenge_overview}.

\begin{table}[ht]
	\centering
	\caption{Instance complexities used for evaluation.}
	\label{tab:challenge_overview}
	\begin{tabular}{|c|c|c|c|c|}
		\hline
		\textbf{Test} & \textbf{Trains} & \textbf{Map size} & \textbf{Cities} & $\boldsymbol{\pi}$\textbf{ Malf.} \\ \hline
		0 & 7 & $30\times30$ & 2 & 1/540 \\\hline
		1 & 10 & $30\times30$ & 2 & 1/900 \\\hline
		2 & 20 & $30\times30$ & 3 & 1/1,800 \\\hline
		3 & 50 & $30\times35$ & 3 & 1/4,500 \\\hline
		4 & 80 & $35\times30$ & 5 & 1/7,200 \\
		\hline
	\end{tabular}
\end{table}

\subsection*{Agent Architecture}
\begin{figure}[hb!]
	\centering
	\includegraphics[width=.95\columnwidth]{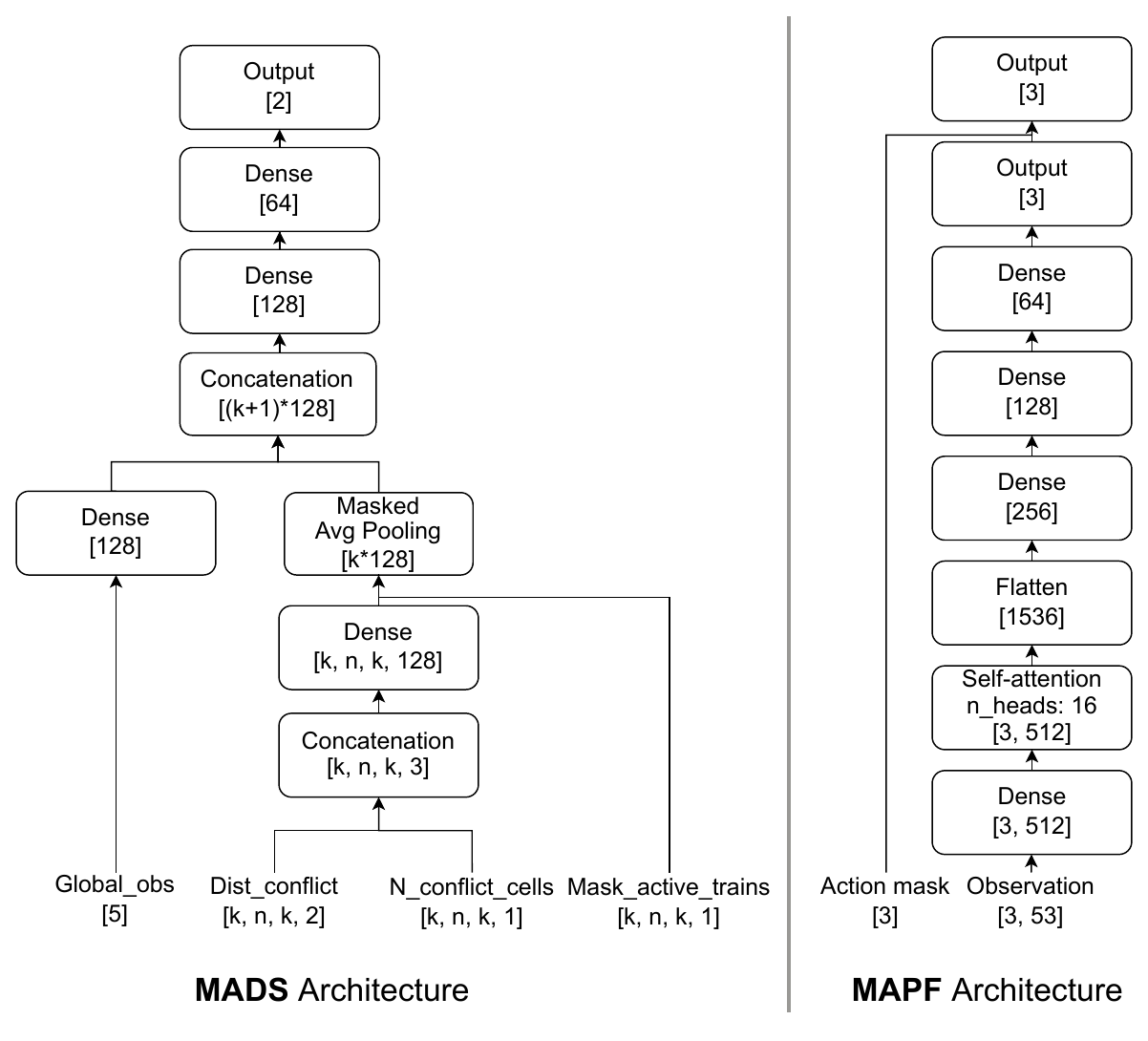}
	\caption{Semi-hierarchical agent architecture.}
	\label{fig:agent-architecture}
\end{figure}

Figure~\ref{fig:agent-architecture} depicts the shared architecture for both \ac{mapf} and \ac{mads}.

For \ac{mads}, input comprises four blocks: a 5-element global‐feature tuple, the \textit{Dist\_conflict} vector, the \textit{N\_conflict\_cells} (depending on active‐train count $n$), and top-$k$ path comparisons, plus a binary mask identifying active trains at each timestep. We fix $n$ to enable mini-batch learning and use the mask to indicate active trains. \textit{Dist\_conflict} and \textit{N\_conflict\_cells} are embedded from $3$ dimensions to $128$, and an average pooling across the masked $n$ trains and their \manuelfeed{top-$k$} paths yields a $k\times128$ vector. The pooling compresses the features associated with the top-$k$ candidate paths of the off-map train, yielding a single summary vector used for the dispatch decision.

Concurrently, \textit{Global\_obs} is embedded to $128$-dim and concatenated with the pooled conflict features. The combined vector passes through dense layers to produce a binary output. Masking is redundant due to binary action space and the \textit{Decision Controller}.
\manuelfeed{All experiments use $k=2$ (top-2 shortest paths). Increasing $k$ offered negligible benefits while substantially enlarging the observation, since the environment must compute path conflicts against all $k$ alternatives for each on-map train.}

For \ac{mapf}, the input is a $[3,53]$ tensor (three $53$-dimensional vectors: the current train plus two candidates for conflicts), with an action mask to block infeasible moves. Each $53$-dim vector is linearly projected to $512$ features, \adrianfeed{processed through a 16-head self-attention module equivalent to a single Transformer encoder layer,} added back to its input and passed through a 10\% dropout (disabled at inference). The output is concatenated into a vector of size $1,536$, which is finally run through three dense layers and combined with the mask.

\christianfeed{Model hyperparameters are selected through a grid search: \ac{mads} uses \textit{Tanh}, while \ac{mapf} uses \textit{ReLU} activations.}

\subsection*{Training}
While specific implementation details are documented in the repository, this section summarises the training process.

\ac{mads} and \ac{mapf} are trained independently through an iterative \ac{bc} procedure. Trajectories are generated using a fixed-budget \ac{mcts} planner following~\cite{elmanyari2025morl}. The search uses policy logits as priors and Monte-Carlo returns as value estimates, with depth limits and simulation budgets chosen to keep rollout time tractable in Flatland-RL. \ac{bc} updates are performed with minibatched training and early stopping based on validation loss. As \ac{bc} quality depends on trajectory quality rather than their origin, we pre-filter the collected data by discarding samples from trains that fail to reach their destination.

Training for \ac{mapf} follows a curriculum that increases network complexity and train density. The refinement cycle is terminated once the agent reaches a satisfactory performance threshold. We stop training once the success rate exceeds \emph{50\% for 15 trains on a $30\times30$ map with 3 cities}, after which additional experience provides diminishing returns. The resulting \ac{mapf} policy is then frozen.

Next, \ac{mads} is trained with the same \ac{mcts}/\ac{bc} loop, but its initialisation requires extra care because dispatch decisions are sparse. We first generate a dataset using a simple heuristic dispatcher that delays any train whose top-$k$ paths indicate an immediate conflict, as defined in the observation space. This provides clean examples of safe dispatch behaviour, from which an initial \ac{mads} policy is learned via \ac{bc}. We then collect a heterogeneous dataset by running scenarios with 10, 20, and 50 trains, exposing \ac{mads} to a wider range of congestion patterns. The combined dataset is used for refinement through additional \ac{mcts} rollouts and \ac{bc} updates, improving generalisation across densities and map configurations.

During evaluation and rollout, the hierarchical model is reinstated across all agents, with each agent holding an independent copy of the two policies. Performance is measured using the metrics of Table~\ref{tab:metrics}, capturing both solution quality and scalability.

\begin{table}[ht]
	\small
	\centering
	\caption{Evaluation metrics used in experiments.}
	\label{tab:metrics}
	\begin{tabular}{|l|c|c|}
		\hline
		\textbf{Metric name} & \textbf{Interval} & \textbf{Notes} \\ \hline
		Success \manuelfeed{rate}                        & [0, 1] & Higher is better    \\ \hline
		Deadlock \manuelfeed{rate}                       & [0, 1] & Lower is better     \\ \hline
		Arrival delay   & [0, $T_{\max}$] & Lower is better   \\ \hline
		Cancelled \manuelfeed{rate} & [0, 1] & Lower is better     \\ \hline
	\end{tabular}
\end{table}

Evaluation metrics include the \manuelfeed{rate} of trains that successfully reach their target stations (\textit{Success \manuelfeed{rate}}), the proportion of trains permanently blocked by circular resource dependencies (\textit{Deadlock \manuelfeed{rate}}), and the average delay between scheduled and actual arrival times (\textit{Arrival delay}). \adrianfeed{We also report the portion of trains that remain undispatched by the end of the episode (\textit{Cancelled \manuelfeed{rate}}), either because the agent chose not to dispatch them or due to a lack of available resources.} \christianfeed{The maximum possible delay is bounded by the episode length~\(T_{\max}\).}

\section{Results}
\label{sec:results}

\subsection{Speed Impact in Flatland-RL}

\label{ss:speed_consideration}

Before evaluating performance, \manuelfeed{we analysed how fractional and constant speed affect timetable generation in Flatland-RL, with constant-speed trains advancing one cell per timestep and fractional-speed trains moving only every few timesteps.}

While fractional speeds were framed in the challenge documentation as an added source of difficulty, our results suggest the opposite. \christianfeed{Fractional speeds reduce motion per timestep of certain trains, increasing the overall episode horizon. This wider temporal window spreads departures and arrivals, reducing concurrency and resource conflict. In contrast, constant-speed profiles shorten the episodes, significantly increasing congestion.} \christianfeed{To quantify this effect, we reproduced the 10 seeds of {Test~4} from the challenge setup~\cite{laurent2021flatland} and evaluated the operational window, defined as the time difference between the {latest arrival} and the {earliest departure}.}

\begin{figure}[hbt!]
	\centering
	\includegraphics[width=.85\columnwidth]{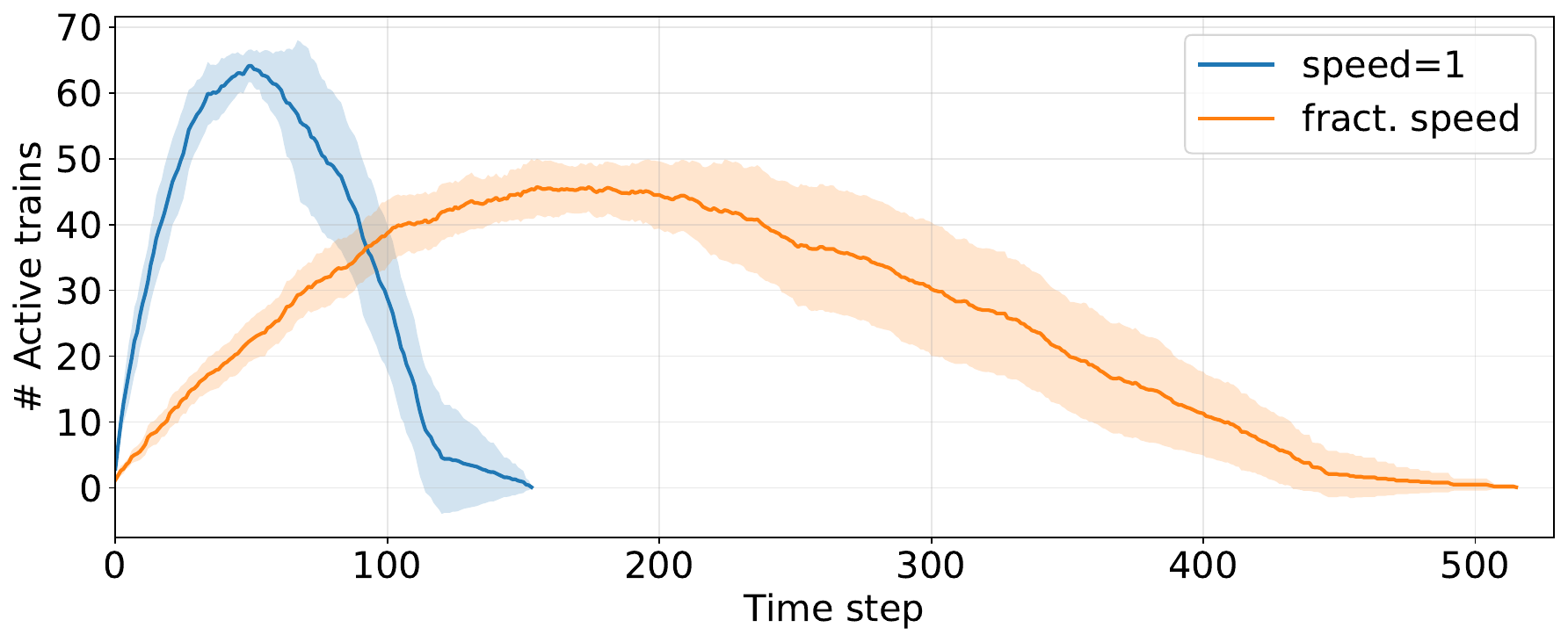}
	\caption{Average active-train concurrency over time for 80 trains over 10 episodes on Test~4 (95\% Confidence Interval).}
	\label{fig:comparison}
\end{figure}

\christianfeed{Figure~\ref{fig:comparison} shows the distribution of active trains, meaning the number of trains simultaneously within their operational window. Constant speed (speed=1) produces much shorter episodes and pushes concurrency to about 80\% of all trains (around 64 active at once), creating strong competition for limited resources. Fractional speeds (fract. speed) lengthen the episode horizon, reducing peak simultaneous activity to roughly 46 trains and easing congestion across the network.}

\subsection{Performance Evaluation of Maze-Flatland}	
To benchmark the hierarchical approach presented in this paper we compare Maze-Flatland against four different baselines: (i) \textit{Greedy}, a distance-minimising agent that selects locally optimal actions; (ii) \textit{\ac{pp}}, the \textit{An\_Old\_Driver}'s method~\cite{li2021scalable}, a heuristic planner that computes collision-free paths through sequential prioritisation; \adrianfeed{(iii) \textit{Deadlock Avoidance}\footnote{Provided by the Flatland Association at: \url{https://github.com/flatland-association/flatland-baselines}.}, a heuristic method that generates collision-free paths with minimal delay and serves as a strong benchmark for fixed-path scenarios without rerouting;} and (iv) \textit{TreeLSTM}~\cite{jiang2022multi}, an \ac{rl} method that achieves competitive performance relative to \ac{or} baselines. Each method is evaluated over 50 independent episodes across the test levels defined in Table~\ref{tab:challenge_overview}.

\begin{figure}[hbt!]
		\centering
		\includegraphics[width=\columnwidth]{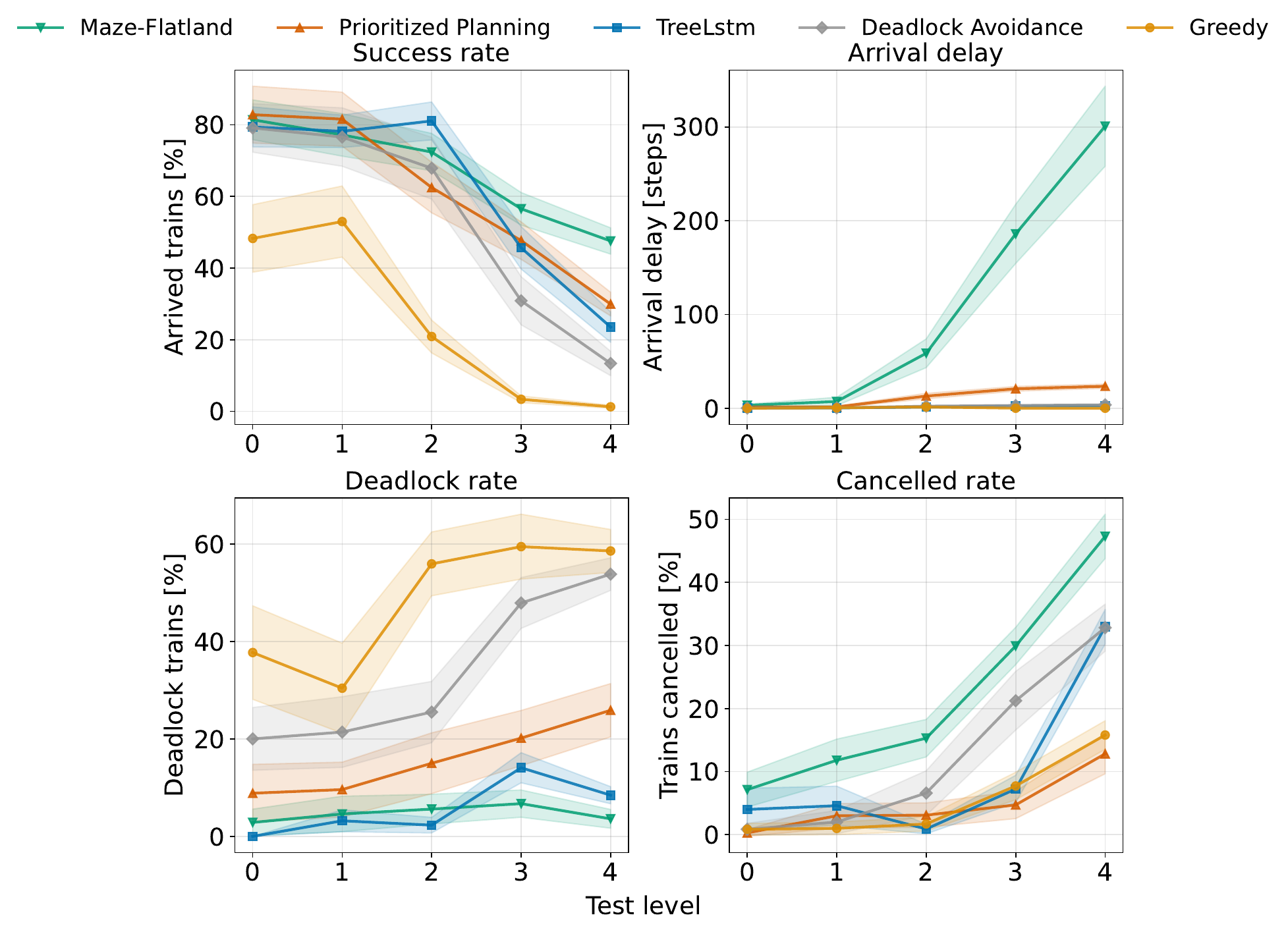}
		\caption{Mean performance over 50 episodes across 5 levels (95\% Confidence Interval).}
		\label{fig:res_full_picture}
	\end{figure}

As shown in Figure~\ref{fig:res_full_picture}, for test levels \textit{0} and \textit{1}, all methods show comparable performance across the evaluated metrics, except for \textit{Greedy}, which suffers from a higher deadlock \manuelfeed{rate}, resulting in a lower success rate. \ac{pp} attains the highest Success \manuelfeed{rate}, successfully dispatching nearly all trains on the map (Cancelled \manuelfeed{rate} is 0\% for \textit{Test~0} with 7 trains and below 5\% for \textit{Test~1} with 10 trains). However, dispatching all trains simultaneously increases the likelihood of deadlocks, where a subset of trains block one another's paths (Deadlock \manuelfeed{rate}).

	As the load increases, Maze-Flatland consistently delivers more trains to their destinations, with about half completing their journeys in the 80-train scenario (\textit{Test~4}). Although it shows a higher cancellation rate under extreme load, Maze-Flatland maintains near deadlock-free operation, averaging no more than four trains stuck in deadlock per episode ($\leq5\%$).
	
	\adrianfeed{Similarly, the \textit{Deadlock Avoidance policy} reduces deadlock occurrence but at the cost of a higher cancellation rate compared to the \textit{Greedy} baseline.} Overall, Maze-Flatland achieves the best trade-off by minimising the average number of trains trapped in deadlock, despite a higher and growing arrival delay as traffic density increases.

	The cancellation rate stems from the conservative behaviour of the MADS policy, which must accommodate as many trains as possible within the limited episode horizon imposed by the Flatland-RL environment and its constant speeds, as discussed in Section~\ref{ss:speed_consideration}. Cancellations are not explicit \ac{mads} decisions; they arise when delayed trains lack time to reach their destinations before the episode ends.

These results confirm that the hierarchical structure of Maze-Flatland enhances robustness and scalability. The \ac{mads} component dynamically delays train departures according to on-map conditions (though it is overly conservative in smaller instances), while the \ac{mapf} component ensures deadlock-free planning and rescheduling under malfunctions. Together, these mechanisms enable the hierarchical design to balance efficiency and stability across varying conditions.
	
	The reproduced results for both baselines differ from those reported in the respective papers for \manuelfeed{\textit{\ac{pp}}~\cite{li2021scalable} and \textit{TreeLSTM}~\cite{jiang2022multi}}. This variation arises \christianfeed{from the different speed profile used at initialisation. This work prioritises maximising the success rate and reducing deadlocks, and therefore employs constant speed profiles, whereas the original baseline implementations were designed and evaluated using fractional speeds.} As noted in Section~\ref{ss:speed_consideration}, the operation time window increases inversely with the maximum train speed. Consequently, the increased competition for shared resources substantially degraded performance. Nevertheless, Maze-Flatland demonstrates the most effective trade-off, maximising the number of trains successfully reaching their destinations.
	
	To further assess the contribution of each module, we perform an ablation study analysing the independent effects of the \ac{mads} and \ac{mapf} components.
	
	\section{Ablation study}
	\label{sec:ablation}
	
\begin{figure}[hbt!]
		\centering
		\includegraphics[width=\columnwidth]{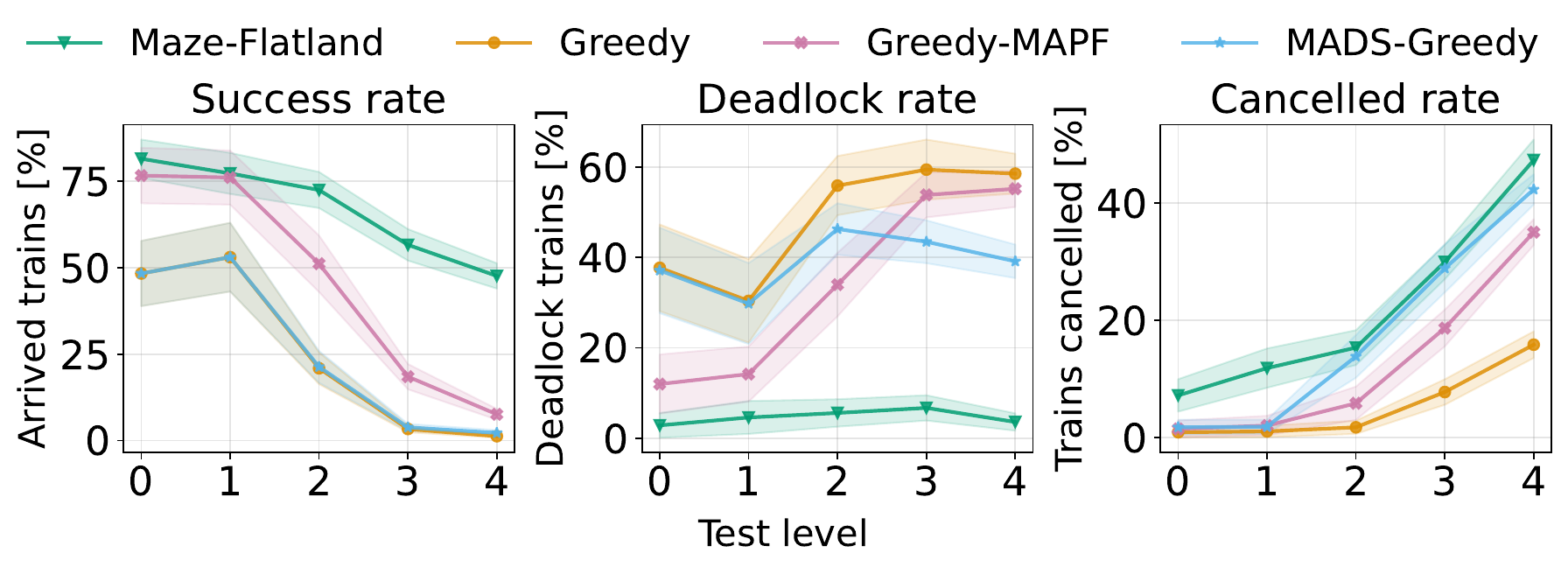}
		\caption{Ablation study showing performance of hybrid configurations combining \ac{mads} and \ac{mapf} with greedy modules.}
		\label{fig:ablation_res}
	\end{figure}

	\adrianfeed{In addition to the Maze-Flatland setups, we evaluate two hybrid configurations combining or replacing the \ac{mads} and \ac{mapf} modules with greedy counterparts to assess their individual contributions to overall performance.} The results, shown in Fig.~\ref{fig:ablation_res}, illustrate the effect of each component across all test levels. The complete Maze-Flatland configuration clearly outperforms both hybrid and greedy variants, particularly under dense traffic conditions.

	The two hybrid configurations perform better than the \manuelfeed{\textit{Greedy}} setup. \textit{\ac{mads}-Greedy}, which combines the intelligent dispatcher with a greedy routing policy, behaves similarly to the greedy baseline but mitigates deadlocks by proactively cancelling trains. Conversely, \textit{Greedy-\ac{mapf}} matches \textit{Maze-Flatland} performance in low-density scenarios (up to 10 trains, \textit{Test~1}), where all trains can typically be dispatched. As traffic increases, however, its performance declines due to congestion, resulting in a deadlock \manuelfeed{rate} comparable to the \manuelfeed{\textit{Greedy}} setup.

	Overall, the ablation confirms that both components are essential for robust performance. The \ac{mads} policy reduces deadlocks through adaptive \christianfeed{departure scheduling}, especially in dense scenarios with strong resource competition, while the \ac{mapf} module maintains efficient routing under congestion. Their joint operation enables the system to preserve stability and throughput as network density increases.
	
	\section{Conclusion}
	\label{sec:conclusion}
	
	\adrianfeed{
		This paper highlights an imbalance in agent behavior during training, where on-map actions dominate while dispatch decisions remain under-explored. The action used for dispatching coincides with the most frequent action in low-density scenarios (up to 10 trains), further amplifying this imbalance. These under-explored state–action pairs strongly influence future performance and may even lead to non-recoverable states, explaining the difficulties faced by \ac{rl} methods.}
	
	To address this issue, we present Maze-Flatland, a semi-hierarchical \ac{marl} framework for the \ac{vrsp} in railway operations. The proposed design separates decision-making into two complementary components: \ac{mads}, which governs decentralised dispatching through adaptive scheduling, and \ac{mapf}, which handles routing, \christianfeed{prioritisation through stopping}, and replanning under congestion and malfunctions.
	
	By disentangling dispatching and routing, the semi-hierarchical approach enables each policy to focus on task-relevant observations and act over a minimal action space, improving coordination, scalability, and stabilising learning.
	
	Experimental results confirmed that the hierarchical structure enhances robustness across different traffic densities. Maze-Flatland outperforms heuristic and monolithic \ac{rl} baselines, achieving higher throughput and maintaining near deadlock-free operation even under heavy load. The ablation study further confirmed the complementary roles of the two modules: \ac{mads} prevents network saturation through controlled dispatching, while \ac{mapf} sustains efficient routing in dense traffic.

The modular design is expected to improve computational efficiency by reducing the cost of exploration through action-space dimensionality reduction.

Future work will optimise Maze-Flatland for scenarios with fractional train speeds to verify performance trends remain consistent under variable dynamics. Additionally, current agents prioritise deadlock-free operation and maximising arrival rate over minimising delays. We plan to explore multi-objective optimisation strategies to jointly optimise delay, throughput, and resource utilisation, bringing the framework closer to real-world railway operations.

\section*{Acknowledgments}

\makeack

\bibliographystyle{IEEEtran}
\bibliography{sample.bib}

@misc{ecoffet1901go,
  title={Go-explore: A new approach for hard-exploration problems. arXiv. 2019},
  author={Ecoffet, Adrien and Huizinga, Joost and Lehman, Joel and Stanley, Kenneth O and Clune, Jeff},
  year={1901}
}

@article{jiang2022multi,
  title={Multi-agent path finding via tree lstm},
  author={Jiang, Yuhao and Zhang, Kunjie and Li, Qimai and Chen, Jiaxin and Zhu, Xiaolong},
  journal={arXiv preprint arXiv:2210.12933},
  year={2022}
}

@article{lindenmaier2022efficient,
  title={Efficient Real-time Rail Traffic Optimization: Decomposition of Rerouting, Reordering, and Rescheduling Problem},
  author={Lindenmaier, L{\'a}szl{\'o} and L{\"o}v{\'e}tei, Istv{\'a}n Ferenc and Aradi, Szil{\'a}rd},
  journal={arXiv preprint arXiv:2209.12689},
  year={2022}
}

@inproceedings{roost2020improving,
  title={Improving sample efficiency and multi-agent communication in RL-based train rescheduling},
  author={Roost, Dano and Meier, Ralph and Huschauer, Stephan and Nygren, Erik and Egli, Adrian and Weiler, Andreas and Stadelmann, Thilo},
  booktitle={2020 7th Swiss Conference on Data Science (SDS)},
  pages={63--64},
  year={2020},
  organization={IEEE}
}

@inproceedings{laurent2021flatland,
  title={Flatland competition 2020: MAPF and MARL for efficient train coordination on a grid world},
  author={Laurent, Florian and Schneider, Manuel and Scheller, Christian and Watson, Jeremy and Li, Jiaoyang and Chen, Zhe and Zheng, Yi and Chan, Shao-Hung and Makhnev, Konstantin and Svidchenko, Oleg and others},
  booktitle={NeurIPS 2020 Competition and Demonstration Track},
  pages={275--301},
  year={2021},
  organization={PMLR}
}

@inproceedings{silver2005cooperative,
  title={Cooperative pathfinding},
  author={Silver, David},
  booktitle={Proceedings of the aaai conference on artificial intelligence and interactive digital entertainment},
  volume={1},
  pages={117--122},
  year={2005}
}

@inproceedings{li2021scalable,
  title={Scalable rail planning and replanning: Winning the 2020 flatland challenge},
  author={Li, Jiaoyang and Chen, Zhe and Zheng, Yi and Chan, Shao-Hung and Harabor, Daniel and Stuckey, Peter J and Ma, Hang and Koenig, Sven},
  booktitle={Proceedings of the international conference on automated planning and scheduling},
  volume={31},
  pages={477--485},
  year={2021}
}

@inproceedings{ma2017multi,
  title={Multi-agent path finding with delay probabilities},
  author={Ma, Hang and Kumar, TK Satish and Koenig, Sven},
  booktitle={Proceedings of the AAAI Conference on Artificial Intelligence},
  volume={31},
  year={2017}
}

@inproceedings{shaw1998using,
  title={Using constraint programming and local search methods to solve vehicle routing problems},
  author={Shaw, Paul},
  booktitle={International conference on principles and practice of constraint programming},
  pages={417--431},
  year={1998},
  organization={Springer}
}

@inproceedings{schneider2024intelligent,
  title={Intelligent Railway Capacity and Traffic Management Using Multi-Agent Deep Reinforcement Learning},
  author={Schneider, Stefan and Ramesh, Anirudha and Roets, Anne and Stirbu, Ciprian and Safaei, Farhad and Ghriss, Faten and W{\"u}lfing, Jan and G{\"u}ra, Mehmet and Sibon, Nima and Gentry, Rick and others},
  booktitle={2024 IEEE 27th International Conference on Intelligent Transportation Systems (ITSC)},
  pages={1743--1748},
  year={2024},
  organization={IEEE}
}

@article{kaelbling1998planning,
  title={Planning and acting in partially observable stochastic domains},
  author={Kaelbling, Leslie Pack and Littman, Michael L and Cassandra, Anthony R},
  journal={Artificial intelligence},
  volume={101},
  number={1-2},
  pages={99--134},
  year={1998},
  publisher={Elsevier}
}

@inproceedings{zhang2024improving,
  title={Improving the efficiency and efficacy of multi-agent reinforcement learning on complex railway networks with a local-critic approach},
  author={Zhang, Yuan and Deekshith, Umashankar and Wang, Jianhong and Boedecker, Joschka},
  booktitle={Proceedings of the International Conference on Automated Planning and Scheduling},
  volume={34},
  pages={698--706},
  year={2024}
}

@misc{flatland-git,
  author       = {{Flatland Association}},
  title        = {flatland-rl: OpenAI Gym Environment for Railway Management},
  howpublished = {\url{https://github.com/flatland-association/flatland-rl}},
  year         = {2025},
  note         = {GitHub repository, accessed 2025-07-22}
}

@article{dietterich2000hierarchical,
  title={Hierarchical reinforcement learning with the MAXQ value function decomposition},
  author={Dietterich, Thomas G},
  journal={Journal of artificial intelligence research},
  volume={13},
  pages={227--303},
  year={2000}
}

@book{ferber1999multi,
  title={Multi-agent systems: an introduction to distributed artificial intelligence},
  author={Ferber, Jacques and Weiss, Gerhard},
  volume={1},
  year={1999},
  publisher={Addison-wesley Reading}
}

@inproceedings{elmanyari2025morl,
author = {El Manyari, Yassine and Fuxjaeger, Anton Robert and Zahlner, Stefan and van Dijk, Joost and Castagna, Alberto and Barbieri, Davide and Viebahn, Jan and Wasserer, Marcel},
title = {Efficient Multi-Objective Optimisation for Real-World Power Grid Topology Control},
year = {2025},
isbn = {9798400711251},
publisher = {Association for Computing Machinery},
address = {New York, NY, USA},
url = {https://doi.org/10.1145/3679240.3734659},
doi = {10.1145/3679240.3734659},
booktitle = {Proceedings of the 16th ACM International Conference on Future and Sustainable Energy Systems},
pages = {614–621},
numpages = {8},
location = {
},
series = {E-Energy '25}
}

@article{li2007vehicle,
	title={The vehicle rescheduling problem: Model and algorithms},
	author={Li, Jing-Quan and Mirchandani, Pitu B and Borenstein, Denis},
	journal={Networks: An International Journal},
	volume={50},
	number={3},
	pages={211--229},
	year={2007},
	publisher={Wiley Online Library}
}

\end{document}